\documentclass{article}

    \PassOptionsToPackage{numbers, compress}{natbib}

    \usepackage[final]{neurips_2020}

\usepackage[utf8]{inputenc} 
\usepackage[T1]{fontenc}    
\usepackage{hyperref}       
\usepackage{url}            
\usepackage{booktabs}       
\usepackage{amsfonts}       
\usepackage{nicefrac}       
\usepackage{microtype}      

\usepackage{graphicx}
\usepackage{subfigure}
\usepackage{booktabs} 
\usepackage[switch, modulo]{lineno}
\title{Active Hybrid Classification}

\author{%
  Evgeny Krivosheev\\
  University of Trento, Italy\\
  \texttt{evgeny.krivosheev} \\
  \texttt{@unitn.it}
  \And
    Fabio Casati\\
  Servicenow, USA\\
  \texttt{fabio.casati} \\
  \texttt{@servicenow.com}
  \And
    Alessandro Bozzon\\
      TU Delft, The Netherlands\\
  \texttt{a.bozzon}\\
  \texttt{@tudelft.nl}
}
\begin{document}

\maketitle
\begin{abstract}
Hybrid crowd-machine classifiers can achieve superior performance by combining the cost-effectiveness of automatic classification with the accuracy of human judgment. 
This paper shows how  crowd and machines can  support each other in tackling classification problems. Specifically, we propose an architecture that orchestrates \emph{active learning} and \emph{crowd classification} and combines them in a virtuous cycle. 
We show that when the pool of items to classify is finite we face a learning vs. exploitation trade-off in hybrid classification, as we need to balance crowd tasks optimized for creating a training dataset with tasks optimized for classifying items in the pool.
We define the problem, propose a set of heuristics and evaluate the approach on three real-world datasets with different characteristics in terms of machine and crowd classification performance,  showing that our active hybrid approach significantly outperforms baselines.
\end{abstract}

\section{Introduction and Motivation}

In this paper we focus on \textit{screening} problems where we look for items satisfying a conjunction $p_1 \wedge p_2 \wedge \dots p_n$ of predicates within a \textit{finite pool} of candidate items. 
Screening problems occur frequently in practice  \cite{Mortensen2016crowd,Krivosheev_www18,CrowdDB2011,Lan_dynamicfilter_hcomp17}. 
For example, we may want to look
for \textit{kids-friendly} hotels \textit{close to subway stations} given a pool of hotel descriptions, or
for scientific papers that measure \textit{happiness} in \textit{older adults} given a set of candidate papers obtained via keyword search.
Arguably, no currently available search engine is able to answer such queries with generally applicable approaches (e.g. keyword search). Unless a specific screening logic based on domain-specific knowledge can be implemented, there are essentially two ways to screen items of interest: i) train and exploit machine learning (ML) algorithms or ii) resort to human classification (via experts or by crowdsourcing).


Research in crowdsourcing has made impressive progress in the last few years, and the crowd has been shown to perform well even in difficult tasks \cite{Law_hearth_cscw18,Ranard_harnessing_2014}.  
However, crowdsourcing remains expensive, especially when aiming at high levels of accuracy, which often implies 
collecting more votes per item to make classification more robust to workers' errors.
ML algorithms have also been rapidly improving.
However, in many cases they cannot yet deliver the required levels of precision and recall, typically due to a combination of difficulty of the problem and (lack of) availability of a sufficiently large dataset.
Consider for example the problem of screening scientific literature: ML works very well for retrieving papers that describe \textit{randomized controlled trials} (RCT) \cite{wallaceRct}, also thanks to the availability of very large training datasets.
Instead, when looking for papers measuring happiness among older adults, ML approaches are  challenged both by the lack of training data
and by the intrinsic difficulty of the problem:  papers may not even mention the term \textit{happiness} but only related constructs, or again happiness may be discussed but not measured.

\textit{Hybrid classifiers} (HCs) are rapidly emerging as a way for ML and crowds to join forces to improve accuracy and reduce costs beyond the traditional process of labeling data, training a model, and then using a model to classify items automatically \cite{imran2016twitter,imran2014aidr}. The underlying idea is that ML algorithms, even when weak for the problem at hand, can still be exploited. 
For instance, they can be used to reduce the workload of the crowd by labeling items for which they have high confidence, 
or they can contribute to the classification decision (e.g., by casting a vote, just like crowd workers do)  \cite{Vaughan_crowd17,Kamar_2012_combining}.  
For example, in medical diagnosis, crowd and machine have been shown to be able to jointly screen and select cases to be brought to the attention of medical doctors \cite{Law_hearth_cscw18} with impressive accuracy.

In this paper we are specifically interested in cases where 
no pre-existing trained ML model is available.
Under such circumstances, \emph{active learning} (AL) techniques can be leveraged to optimize the cost/quality trade-off in the learning process.
However, at some point through the AL process we can decide to stop collecting training data (in a way, stop \textit{learning}) and use the remaining budget to \textit{exploit} the learned model by applying hybrid classification (or simple ML classification if the model accuracy is sufficient).
Furthermore, the \textit{active learning} policy and the \textit{active crowd classification} policy (selecting remaining items in the pool with the goal of classifying them efficiently) may prioritize different items.
In the following we refer to the balancing of active learning and active classification in finite pool contexts as \textit{active hybrid classification} (AHC).
What makes the problem challenging is that when we approach a new task, in general we do not know how well ML or crowd can perform and what is the learning curve of the given ML classifiers. The available budget and pool size (which are instead typically known) also come into play when deciding an AHC policy.


\noindent\textbf{Problem Formulation}. 
Given a crowdsourcing task, a set of ML algorithms (which we assume to be initially untrained, for a specific classification problem) with their \emph{active learning} policy, and a crowd classification algorithm with its \emph{active classification} policy, our goal is to identify how to balance the learning versus exploitation trade-off to improve the overall classification cost and quality.

\noindent\textbf{Original Contribution}. In this paper we contribute a formulation of active hybrid classification. 
We propose a generic approach for AHC (independent of the specific ML or HC algorithm adopted) and show that striking the right balance between learning and exploitation can have a significant impact on classification cost and quality. We then propose and discuss several budget allocation heuristics that can be adopted to efficiently balance learning and exploitation. Such heuristics are assessed via a set of experiments on datasets with diverse characteristics to establish their validity (or lack thereof) in different contexts, and ultimately derive some rules of thumb of wide applicability.

\section{Related Work}

\noindent \textbf{Hybrid Machine/Crowd Classification.}
Hybrid classification has become an active area of research in recent year (see  \cite{Vaughan_crowd17} for a survey), drawing particular interest in the context of clustering or entity resolution (\cite{crowdER_2012,VesdapuntER_vldb2014,Vinayak_crowdcluster_nips2016,Gomes_crowdclustering_nips2011}).  
One of the most interesting contribution in crowdsourced classification comes from Kamar et al~\cite{Kamar_2012_combining}, who leverage \textit{crowd} features (such as votes, work completion times) and \textit{task} features to learn a model. 
Bernstein and colleagues also turn to the crowd to learn a model but do so by explicitly asking the crowd to extract features and patterns to then be leveraged by classifiers ~\cite{flock_2015,Carlos_pattern}.
These results are complementary to our work: We do not aim at identifying ``good'' features or effective ML algorithms, which is an orthogonal problem. 
We treat ML algorithms as black boxes and use them to improve  classification cost or quality.
Other HC approaches are based on first attempting to classify items via ML, and then resorting to human classification on items for which ML has low confidence \cite{mozafari2014scaling,Law_hearth_cscw18}, or by leveraging ML to structure the crowdsourcing task (e.\,g., clustering items to optimally match them with workers \cite{VesdapuntER_vldb2014,Vinayak_crowdcluster_nips2016}), or
 by leveraging the ML output to compute a prior belief on each item
\cite{Krivosheev_cscw18}.
This belief informs their crowd classification strategy by progressively adjusting the number of votes requested on each item based on whether the crowd confirms or negates such belief. No prior assumption is made on ML classifiers accuracy, and the hybrid algorithms are designed to be robust to weak classifiers. 
Finally, Nguyen et al. \cite{Nguyen2015}  propose to leverage ML in crowd classification to determine i) which item to collect votes for and ii) whether to do so with crowd workers or with experts, assumed to be more expensive but more accurate.



\noindent \textbf{Active Learning.}
\textit{Active Learning} (AL) studies how to select items to improve the learning rate of  ML classifiers with respect to random selection \cite{Aggarwal_2014_AL}.
One of the most popular active learning strategies, dating back to the 1990s, is \textit{uncertainty sampling} \cite{Lewis_1994_uncertainty}, where manual labeling is sought first on items for which the classifier is less certain. 
Mozafari \cite{mozafari2014scaling} proposes to also consider items that have the biggest impact on the learned model if the classifier is found to be wrong in the prediction.
Interestingly, this paper also discusses hybrid classification and the issue of when to stop collecting votes to train the classifier. However, it deals neither with finite pool problems where there is a trade off between learning and exploitation, nor with screening problems. 
Abundant research has since proposed a number of different approaches and optimizations, and recent studies have shown that active learning can be successfully exploited  in deep networks as well \cite{al_dn_W17-2630}.
Although the problem we deal with here has aspects of active learning - since we also identify how to prioritize items to label, our goal is to balance learning vs exploitation, and the item prioritization is then made by the ML or HC algorithms based on their own logic.
This being said, we do show in the experiments and analysis section how the choice of specific AL policies affect the performance of hybrid classifiers.

\noindent \textbf{Learning to Learn.} 
The problem of balancing learning and exploitation in HC contexts can be cast as a \textit{multi-armed bandit} (MAB) problem (see \cite{Lattimore_book_2019} for a wonderful introduction to MAB).
Baram et al \cite{COMB_MAB_Baram_2004} study how, given an ensemble of AL algorithms, we can choose the items on which to obtain gold labels for training, reassessing the decision after a round of N items.
The problem is mapped to an \textit{adversarial}, contextual MAB problem where the items to be selected are the \textit{arms} of the bandit and the different AL algorithms are the \textit{experts}, each of them indicating  a different set of items (for example, uncertainty sampling vs random sampling). 
They empirically show that by extending \textit{Exp4} \cite{auer2002} - a popular algorithm for adversarial MAB - with a method for estimating the AL reward they can identify the best AL algorithm in the ensemble for different datasets - and even outperform the single best algorithm. 
Hsu and Lin \cite{Hsu_AAA_2015} extend this result following a similar approach but leveraging an improvement of Exp4, called \textit{Exp4.P}, which achieves regret bounds with high probability \cite{pmlr-v15-beygelzimer11a}. 
They also adopt a reward function to measure the performance of the trained classifiers based on importance-weighted accuracy \cite{pmlr-v22-ganti12}, and show that results often outperform previous approaches.

Our problem is similar in that we have to choose between alternatives in an explore vs exploit fashion typical of MAB problems, and as in the research above we cannot assume i.i.d. rewards for arms.
However,  we observe that we do have expectations on the specific shape of ML and CC curves which neither give i.i.d nor adversarial rewards in the strict sense, but tend to follow learning/exploitation curves, although noisy ones and with local minima. We can therefore exploit this knowledge.
Furthermore, we cannot adopt the notion of rewards as suggested in the literature above, both because we do not  deal only with AL arms and, even for the AL side, the contribution of the additional training is not in terms of increased accuracy but rather on how this increased accuracy impact the crowd classification cost/quality trade-off for the items left to classify in our finite pool.



\section{Method and Problem Formulation}

\begin{figure}[thb]
    \centering
     \includegraphics[width=1.0\textwidth]{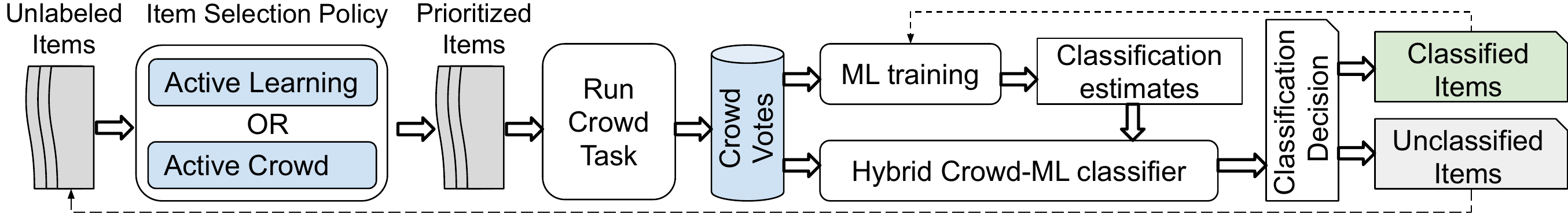}
    \caption{Architecture of the Active Hybrid Classification.}
    \label{fig:arch}
\end{figure}

Figure \ref{fig:arch} shows the high level architecture of the proposed  Active Hybrid Classifier (AHC). AHCs processes items in iterations consisting of three phases: i) selecting a batch of items and predicates to submit to the crowd for voting, ii) obtaining the votes (performing the actual crowdsourcing task), and iii) processing the votes by first re-training ML classifiers with the new information and by then leveraging the ML contribution to guide the crowd classifier. 
Batches are processed until either all items have been classified or the budget has been exhausted - at which point items for which no decision could be taken are left for expert or ML-only classification.
Interestingly, we face the same item (and predicate) selection problem in crowdsourced classification as in Active learning. We refer to the problem of item (and predicate) prioritization for crowd classification as \textit{active crowd classification}, for symmetry with active learning. 
To see this, consider the hotel search example: humans can easily screen hotels based on location by glancing over a map, thereby filtering out the majority of items and leaving a much smaller number of hotels on which to assess the more complex \textit{kids-friendly} predicate. 
Therefore, we can choose to first poll the crowd only on location. Then, for those hotels that pass the location screening, we can obtain votes for kids-friendliness. 
This approach can be adopted in any screening problem, and is followed in  query optimization as well \cite{Hellerstein93predicates}. The challenge in this scenario is that \textit{active learning and crowd active classification are often at odds}: they can select different items and predicates for crowd voting, e.\,g., active learning can select the closest items to the decision boundary while crowd active classification can select easiest items to classify with a few number of crowd votes.
Indeed, the active classification approach based on predicate selectivity mentioned above is particularly ``bad'' for training ML algorithms: not only it does not follow active learning practices such as uncertainty sampling, but, by focusing on eliminating items efficiently, it creates an extremely unbalanced dataset. 
This means that we may be 
facing a \textit{learning} vs \textit{exploitation} trade-off, i.e., whether we invest budget to train ML efficiently (selecting items via \textit{active learning}), or to exploit the trained algorithms and classify items in our pool (selecting items via \textit{active crowd classification} algorithms and using HC to classify).
Intuitively, if the pool is small and the problem is hard for ML, it may be convenient to lean towards exploitation. If the problem is easy for ML and the pool is large, focusing more on learning may be the best strategy.




\noindent \textbf{Problem Formulation.}
The problem can be formulated follows.
\textit{Given a tuple} $(I,PR,A,B,Q)$ \textit{consisting of:}
i) a finite pool of items $I$ to be screened based on a conjunction of predicates $PR = (p_1 \wedge p_2 \dots \wedge p_n)$, 
ii) a set of algorithms $A=(A_{ml},A_{al},A_{hc}, A_{ac})$ which includes a ML classifier $A_{ml}$, an active learning algorithm $A_{al}$, a hybrid classification algorithm $A_{hc}$, and an \textit{active crowd classification} algorithm $A_{ac}$.
iii) a quality/cost trade-off goal $Q$, and a budget $B$.
We aim to identify a decision strategy that defines how to balance \emph{learning} -- i.e., selecting items with the goal of improving $A_{ml}$ efficiently, thanks to active learning $A_{al}$, versus \emph{exploitation} -- selecting items via $A_{ac}$ with the goal of reaching a classification decision  efficiently, also thanks to the contribution of the (partially trained) classifier  $A_{ml}$. 

We observe that rather than having one ML algorithm for the overall screening problem, it is often convenient to have a separate classifier $A^p_{ml}$ for each predicate $p$.
There are several reasons for this: besides being a requirement of some HC algorithms, having separate classifiers facilitates their reuse \cite{cherman2016active}. 
For example, in literature reviews it is not uncommon to revisit definition of predicates (e.g., we realize that ``older adults'' is ambiguous and we want to be more specific, such as ``adults past their retirement age in their country of residence'') while keeping the others unchanged, or to apply the same predicate to a different review (many reviews may need to focus on older adults). 
Furthermore, crowdsourcing tasks focused on a single predicate may offer better results when the predicates are complex to understand and training is required.
We therefore tackle the general multi-predicate case and use $A_{ml}$ to denote a \textit{set} of ml algorithms $\{A^p_{ml}\}$ in the following.


The goal $Q$  typically involves tackling a cost-accuracy trade-off, setting a threshold on recall, precision, or $F_\beta$ measures (where $\beta$ skews the F-measure to favor precision or recall), or some other cost function \cite{Mortensen2016crowd}.
The precise form in which it is expressed is not particularly relevant to our discussion, and in the following we simply take $F_\beta$ and budget spent as measures.


\noindent \textbf{AHC Approach and Algorithm}
The idea behind the approach is to proceed in iterations, where at each run we go through the phases described in Figure \ref{fig:arch}, having selected active learning or active crowd classification. 
Borrowing notation from research in multi-armed bandits, we use $t=1,2\dots n$ to refer to iterations and $K=(k_{l},k_{e})$ to refer to the possible actions (arms) that can be undertaken at each iteration, which are limited to \emph{learning} (denoted with $k_{l}$, meaning that we select items via active learning $A_{al}$) and \emph{exploitation} (denoted with $k_{e}$, meaning that we select items via active crowd algorithm $A_{ac}$). At each iteration $t$ we observe a context $c_t=(\Psi_t^{ml},\Psi_t^{hc},CI_t,UI_t, b_t)$, where:
$\Psi_t^{ml} = \{ \psi_{t, ml}^{i,p}\}$ are the predictions, expressed as probability of item $i$ satisfies predicate $p$, as estimated by $A_{ml}^p$ trained over the labels $L_t$;
$\Psi_t^{hc} = \{ \psi_{t, hc}^{i,p}\}$ are the predictions, expressed as probability of item $i$ satisfies predicate $p$, as estimated by $A_{hc}^{i, p}$ that takes as input labels $L_t$ and $\Psi_t^{ml}$;
$CI_t$ is the set of items for which  $A_{hc}$ has reached a decision so far (depending on classification thresholds);
$UI_t$ is the set of items on which $A_{hc}$ is still undecided. It is always true that $UI_t \bigcup CI_t = I$;
$b_t$ is the budget used so far, out of a total budget $B$.









Notice that classification decisions are \textit{only} made as a result of applying hybrid classification algorithm $A_{hc}$. 
This does not mean that asking for crowd votes is always required: when the budget is exhausted, for instance, $A_{hc}$ can decide to leave items unclassified or to trust the ML prediction.
The choice here depends on the specific HC algorithm used, and AHC is agnostic to that.
Given this algorithm, we now discuss policies to tackle the learning vs exploitation trade-off.

\section{Approach and Heuristics}
The problem of selecting which algorithm ($A_{al}$ or $A_{ac}$) to adopt at each iteration for prioritizing items can be cast as a multi-armed bandit problem, where we balance exploration of the effectiveness of each arm and then exploit the findings to classify items efficiently.
This is indeed the approach followed in active learning when choosing among different ML classifiers and active learning algorithms to train them. 
The problem is cast as an adversarial bandit with expert advice \cite{COMB_MAB_Baram_2004,auer2002,pmlr-v15-beygelzimer11a}, where the policy defines the probability distribution over arms $[K]$ based on weights associated to such arms.
As an example, adopting the popular Exp3 algorithm \cite{auer2002} and adapting it to our context would entail:

\noindent 1. Defining and estimating the value (\textit{reward}) $\hat{x}_k(t)$ obtained as a result of havign chosen arm $k$ at iteration $t$. In our context a meaningful reward can be the reduction in the expected cost for classifying all items in the pool.

\noindent 2. Assigning weights $w_{k_l}$ and $w_{k_e}$ to the arms as follows: $w_{k}(t) =  w_{k}(t-1) \cdot e^{0.5 \cdot \gamma \cdot \hat{x}_{k}(t) }$ , where $\gamma$ is the probability we should assign to random (uniform) selection among arms, regardless of what the weights say. The weight for the arm that has not been chosen remains unchanged.

\noindent 3. Determining the probability  of choosing arm k (e.\,g., the probability of choosing $A_{ac}$):
\begin{equation}\label{eq:pexp3}
P(k_l)=(1-\gamma) \frac{w_{k_l}(t)} {w_{k_l}(t) + w_{k_e}(t)} + 0.5*\gamma
\end{equation}

This approach also has specific upper bounds in terms of expected \textit{weak regret} (that measures how much worse is our reward compared with the best among the two ``static'' policies that always choose the same arm) and the literature mentioned above shows extensions to achieve a specified regret with high probability.
While we could follow the same approach, there are three important differences between the problem discussed here and the ``learning by learning'' method discussed above.

\emph{First}, computing the reward here is more challenging. We retain all the  mentioned difficulties of estimating the accuracy of $A_{ml}$ in active learning contexts (the accuracy essentially is the reward in active learning) \cite{Nguyen2015,importance_weigh_Beygelzimer_2009}.
In addition, the effect of the incremental learning performance has to be estimated in the context of its contribution to $A_{hc}$, since $A_{ml}$ is not used to classify but to assist in crowd classification. This brings additional noise to the estimation.
Furthermore, such contribution changes over time, because i) as $A_{hc}$ progresses we are left with more difficult items to screen, which means that the population of items changes, and ii) arms are not independent: pulling the \textit{learn} arm affects the reward we get in the future by pulling the \textit{exploitation} arm.
Finally, classes in nearly every screening problem are highly unbalanced in favor of exclusions. 
This makes not only harder to train algorithms, but - most importantly for our estimation problem - makes it hard to assess accuracy in an active learning context unless a large pool of labeled items is available. 

\emph{Second}, the incremental value of adding budget is likely to decrease as the iterations proceed.
In terms of $A_{ml}$, learning curves generally tend to level off. At the beginning, every new example provides new information; as we progresses, it is likely that some of the information brought by the new item is already incorporated in the algorithm.
This however is not a guarantee, and in fact the learning behavior can be erratic.
Interestingly, the same is often true for active classification: algorithms try to greedily screen items out efficiently, focusing first on doing cheaply what it can do cheaply, and leaving items harder to classify for later iterations or even give up on them and leave them to experts.


\emph{Third}, weak regret over the best arm would be a conservative measure of effectiveness: choosing always the same arm (that is, relying only on $A_{ml}$ or only on $A_{hc}$) has been shown to perform \textit{worse} in many cases than hybrid strategies \cite{Krivosheev_cscw18}.
 

For these reasons, while we see research on optimization and theoretical bounds as  beneficial in the long run, in this paper we focus on two key questions: i) whether studying the \emph{learning} vs \emph{exploitation} trade-off in AHC is even a problem worth studying and under which conditions benefits can be observed, and ii) whether it is possible to identify a set of ``rule of thumbs'' and simple heuristics that work well for a variety of problems with different characteristics.  
Specifically, out of the space of all possible policies $\Pi$, we look here into deterministic policies $\pi \in \Pi$ that begin with the \emph{learning} arm, to then switch at a specific point to the \emph{exploitation} arm. 
We identify the following  policies $\pi$:



\noindent \textbf{Baseline}: The simplest approach is to stick to the same arm: $\pi(c_t) = [1,0]$ or $\pi(c_t) = [0,1]$, for any context, where the array returned by the policy denotes the probability of choosing $A_{ml}$ or $A_{hc}$.

\noindent \textbf{Fixed point switch}. The second option identifies a priori a percentage of the items (or of the available budget) to devote to ML training (that is, selecting the ML policy) and then switching to the crowd policy.
This may seem arbitrary, but experiences with datasets revealed that there are heuristics that work well in a variety of cases, such as the 70/30 or 80/20 split between training and test data for classical ML algorithms. 
For example, if we base the decision on setting a percentage $\gamma$ of budget $B$ for learning, then the policy becomes  $\pi(c_t) = [1,0]$ if $B_t \leq \gamma B$, and $\pi(c_t) =[0,1]$ otherwise.
Notice that while with this heuristic we could cast the problem to the well-known stopping or \textit{secretary problem} \cite{bruss2000}, in our case we do not have independent and randomly shuffled ``secretaries'' (rewards follow a trend) and in general all the reward estimation challenges discussed apply here as well. 

\noindent \textbf{Stochastic policies (with fixed probability)}.
In this class of policies we identify a priori a static distribution over arms $[K]$ (that is, defining the probability $P_l$ of choosing to learn vs exploit), and at each time we select a policy based on that probability. This results in policy $\pi(c_t) =[P_l,1-P_l]$.

\noindent \textbf{Adaptive Contextual Policies}. 
Adaptive policies continuously estimate the characteristics of the problem at hand, and specifically the ability for the ML classifier to rapidly improve to provide useful contributions, as well as the effectiveness of crowd classification.
Intuitively, we begin with the learning arm and estimate the accuracy ($F_\beta$) of $A_{ml}$ as a result of selecting the learning arm as per the literature \cite{Nguyen2015,importance_weigh_Beygelzimer_2009}. 
We can then compute the trend in accuracy (percentage difference $\delta(t)$ in accuracy between iteration, possible smoothed via a moving average) and map values of $\delta(t)$ to a probability value $k_{al}$ of choosing the learning arm via a mapping function $m$ (similarly to what logistic regression does with the logistic curve).  This results in policy $\pi(c_t) =[m(\delta(t)),1-m(\delta(t))]$.



\section{Experiments and Analysis}
\setlength\abovecaptionskip{-5pt}


\subsection{Experiment Objectives and Design}
We experiment the approach and heuristics described here with the goal of: 
i) understanding whether selecting specific policies \textit{can} lead to cost and quality trade-offs that are different from that of the obvious baselines (fixed policy),
ii) evaluating our heuristics and identifying if we can derive \textit{rule of thumbs applicable in different contexts}, and 
iii) understanding which aspects of the crowdsourcing problem may impact the effectiveness of given heuristics
We identify three datasets with different characteristics in terms of accuracy that ML or crowd achieve, and in terms of predicate selectivity: 


\noindent\textbf{Amazon Reviews}.
The dataset contains reviews on Amazon products, with information about the \textit{sentiment} expressed in the reviews and the product category \footnote{https://www.cs.jhu.edu/~mdredze/datasets/sentiment/}. 
We  created a 5000-item dataset with labels for two predicates: 1) whether the review is on a book (\texttt{Books}) 2) and whether it is a \texttt{Negative review}. 
The \texttt{Books} predicate has selectivity $ \theta= 0.61$, while \texttt{Negative review} has $\theta = 0.10$. 
If we look for negative reviews on books, therefore, only about $5\%$ of reviews pass the screening. 
With this dataset, the task is relatively easy for the crowd. ML performs well on the Books predicate and less well (with the SVM algorithm we adopted) on sentiment analysis.

\noindent\textbf{Medical Abstracts}. This two-predicate screening dataset is taken from the OHSUMED test collection \cite{joachims1998text}, where $p_1$ is \textit{``Does the paper study Cardiovascular Diseases?"} and $p_2$ is \textit{``Does the paper describe Pathological Conditions, Signs, and Symptoms?"}. 
Selectivity values are $\theta_1=0.18$ and $\theta_1 =0.28$ for $p_1$ and $p_2$ respectively. 
The dataset is highly unbalanced, with $5\%$ of items out of 34387 abstracts passing the screening, and an harder task for ML.

\noindent\textbf{Systematic Literature Review (SLR)}. This dataset mimics the screening phase of the SLR process, where the abstracts returned from a keywords-based search need to be screened by researchers based on specific predicates. 
We manually annotated 825 items for two predicates: 
$p_1$: \textit{``Does the paper describe an experiment targeting older adults"?}
$p_2$: \textit{``Does the paper describe an intervention study?"}  The selectivity is $\theta_1=0.58$ and $\theta_2=0.20$, and the resulting set of relevant items is $17\%$ of the dataset.
This task is hard for the crowd and even harder for ML classifiers, especially for predicate $p_2$ (accuracy at approximately 0.6).

We experiment with ML algorithms (SVM and Logistic Regression) and AL strategies (random and uncertainty sampling, which is the most common AL approach).
On the crowd side we focus on the \textit{Hybrid Shortest Run} (HSR) algorithms for active crowd classification and hybrid classification, as to the best of our knowledge they represent the state of the art in screening problems \cite{Krivosheev_cscw18}.
We then test the various heuristics by training classifiers, crowdsourcing votes, and  performing the AHC algorithm described earlier.
In addition to the studies with crowdsourced votes, we  simulate how  results would change as key characteristics of the problem vary - for example, in problems with very high or low crowd accuracy - to get a better sense for the conditions under which heuristics can be effective.


\setlength\abovecaptionskip{0pt}

\subsection{Crowdsourcing Tasks}
For Amazon and SLR datasets, we ran tasks on the \textit{Appen} crowdsourcing platform to collect crowd votes. For Amazon reviews, we collected votes for 1000 items, with 5 votes per item and predicate, obtaining a total of 10,000 votes.
The task is shown in Figure \ref{fig:amazonresults}(f). Workers were paid 3 cent/review, with hourly pay rate $>$10\$/hour. 
The initial test screening included 4 questions (100\% accuracy required) and with 25\% of honeypots after that. 
Crowd accuracy for both predicates was 0.94.

For SLR abstracts we collected 5-7 votes per paper, limiting workers' contributions to a maximum of 18 judgments to allow for worker diversity. 
A total of 236 workers from 41 countries contributed with 1321 judgments.  
The payment was of 15 cents per vote, which computing the reading time spent on each abstract was above 10USD/ hour.
As discussed, each worker focuses and votes for one predicate only, since understanding predicates is complex and takes time.
Resulting accuracies here were 0.6 for \textit{intervention} predicate and 0.8 for \textit{older adults}. Source code and datasets are available online~\footnote{\scriptsize{https://github.com/Evgeneus/Active-Hybrid-Classificatoin\_MultiPredicate}}

\subsection{Experiment settings, execution and results}
\begin{figure*}[t]
	\centering
		\includegraphics[width=1.\textwidth]{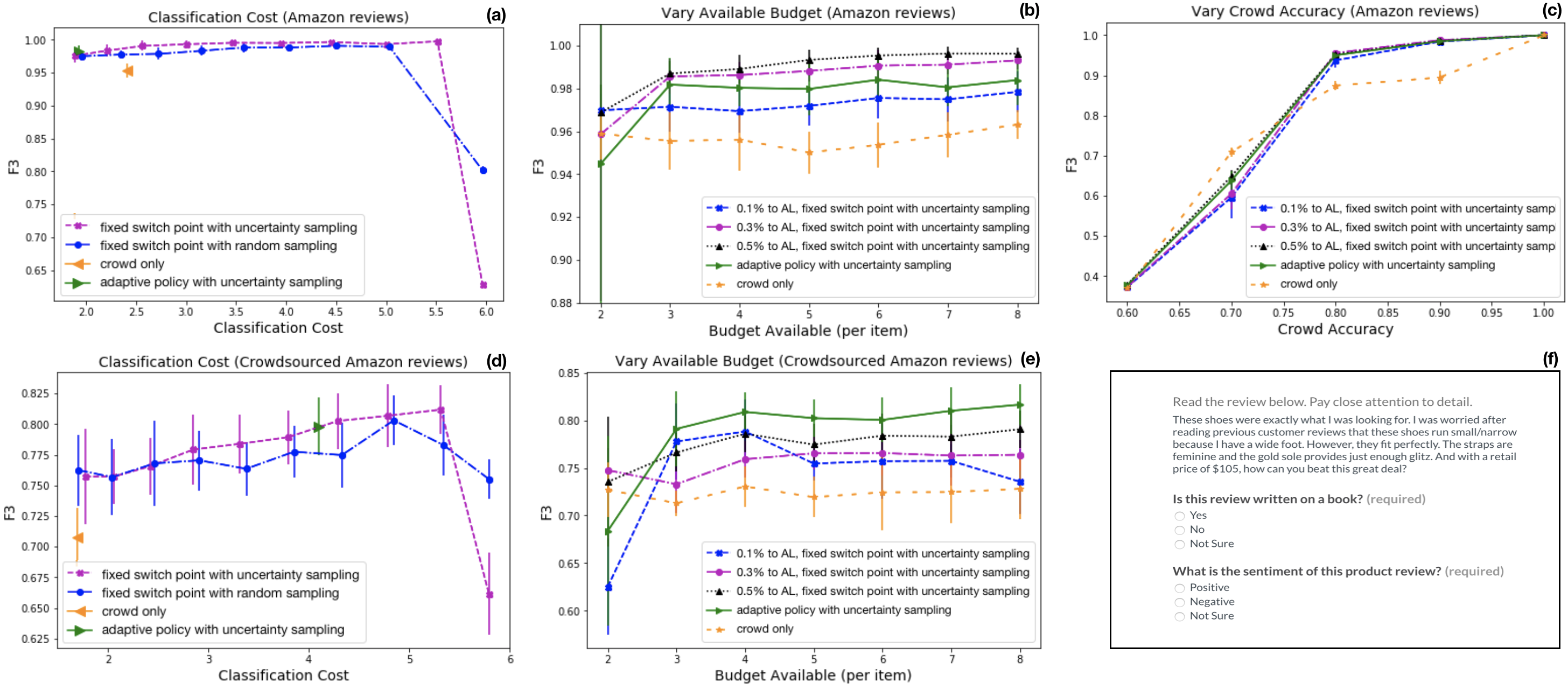}
    		\caption{Results on the Amazon reviews dataset, where \textit{Classification Cost} is the actual cost spent per paper on average (in crowd votes), \textit{Budget Available} is average number of crowd votes we can spend per item. \textit{(f)} shows a screenshot of the Amazon reviews crowdsourcing task on two predicates.}  
\label{fig:amazonresults}
\end{figure*}

To analyze the heuristics, we run AHC over the available datasets, training classifiers and performing hybrid classification first over simulated crowd votes and then on actual votes.

Figure \ref{fig:amazonresults}(a) shows the results of applying the budget allocation heuristics
in terms of classification cost versus $F_\beta$, assuming $\beta=3$ as the dataset are highly imbalanced and we have a few positive items - however the patterns are similar for $\beta=1$ (see the companion material).
AHC was applied to classify a pool of 5000 amazon reviews, and crowd votes were simulated by using the same accuracy values obtained from the crowdsourcing experiments (p1=p2=0.94).
The orange dot corresponds to the static policy that always applies HSR's active crowd classification (no ML training and no hybrid classification - only crowd). 
The purple and blue lines show the baseline and fixed switch point policy for different switch points, expressed as percentage of budget devoted to the learning arm, from 0.1 to 1 (from left to right). The purple curve adopts uncertainty sampling, while the blue denotes random sampling.
The green dot represents an adaptive policy that switches from learning to exploitation when the increase in accuracy falls below 2\% for the first time after the first 500 items.
Vertical bars denote standard deviation over 10 repeated runs of the AHC process, and dots show the mean value. 


\begin{figure*}[htb]
	\centering
    \begin{minipage}{1\linewidth}
		\includegraphics[width=1\textwidth]{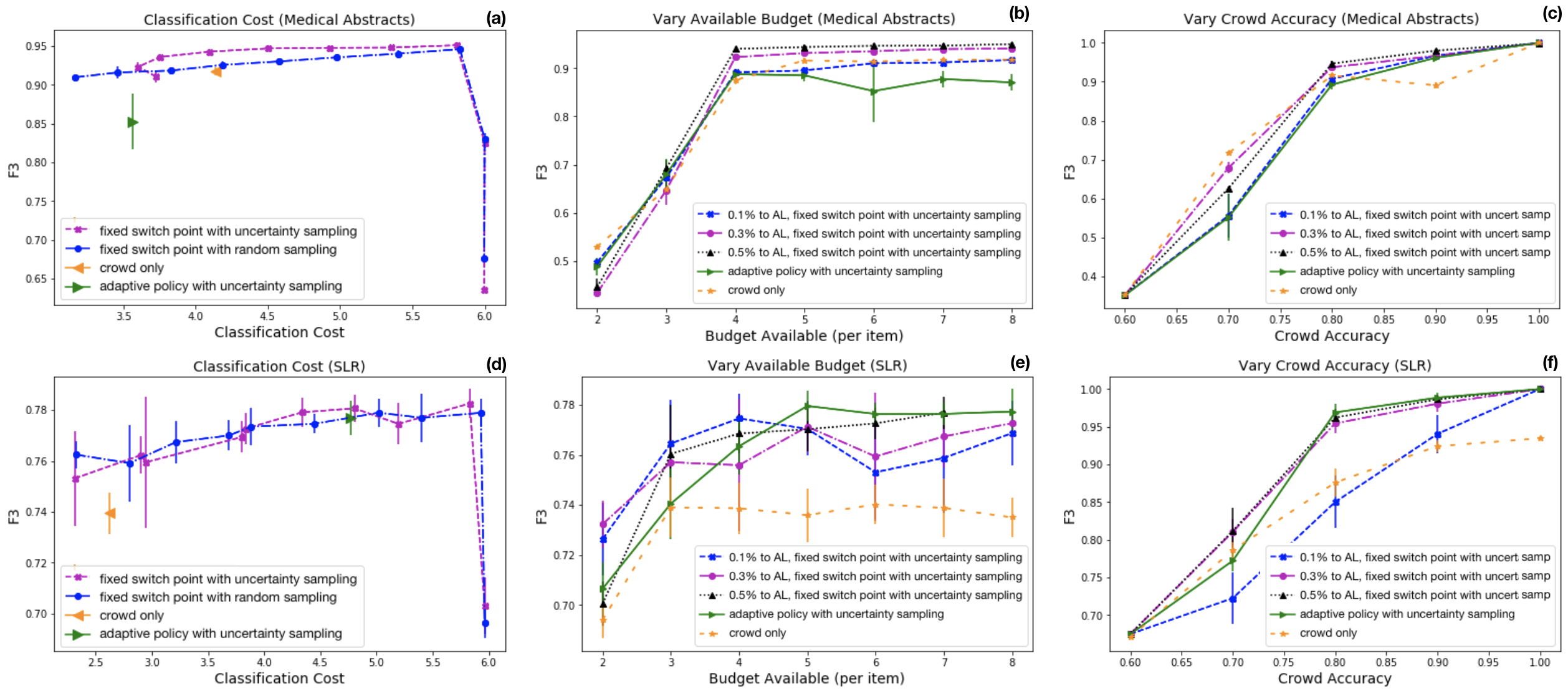}
		
		\caption{Results on the MED Abstracts and SLR datasets with simulated crowd votes.}   
\label{fig:medslr_experiments}
	\end{minipage}
\end{figure*}

The figures show that for small percentages of budget allocated to learning, AHC outperforms crowd-only classification in both quality and cost.
As the learning budget increases, so do both cost and quality so that at some point the choice becomes a trade-off. 
However, when we approach 100\% devoted to learning, the performances drop below that of crowd classification. 
This latter results is in line with work that separately studied performance of machine-only and crowd-only approaches \cite{Krivosheev_cscw18} and confirms that even a ``sprinkle'' of crowd votes is crucial to classification accuracy unless machines are near-perfect. 
Figure \ref{fig:amazonresults}(d) just below shows the same analysis but on crowdsourced data. 
Results are more noisy likely because the crowdsourced dataset is of only 1000 items. With fewer items we also have worse ML accuracy, hence the lower $F_\beta$.

Figures \ref{fig:amazonresults}(b) and (e) show the behavior of the same curves as the overall available budget increases, expressed in average number of votes available per item, for simulated and actual crowd votes respectively. 
Because ML accuracy here is very high, the performances remain good even with a low budget (we will see that this is not the case for other datasets).
Finally, Figure \ref{fig:amazonresults}(c) simulates variations in crowd accuracy.
Notice that the range of values for the $F$ measure here is much wider. 
Here we notice that the benefits of hybrid classification over crowd-only begin to be noticeable when crowd accuracy goes above 0.8, as otherwise the training dataset itself becomes noisy. 



Figures \ref{fig:medslr_experiments}(a) and (d) show the same experiments run on the medical and SLR datasets, where we recall that on the medical datasets we do not have crowdsourced data (we simulated accuracy at random in the [0.6,1] interval), while for the SLR dataset we only have simulated crowd votes and take the real crowd accuracy of 0.6 and 0.8 respectively for the two predicates. 
The trends are similar to the Amazon dataset, and as usual we see the larger variance in Figure (d) as for SLR we only have a small pool of items (just over 800).
We still see however that for small percentages of budget invested in learning the results are pareto-optimal with respect to crowd classification.
Also similar is the impact of crowd accuracy, with the benefit of AHC becoming manifest at crowd accuracies between 0.8 and 1, which are fairly common values (notice that the difference in the F measure at 0.9 is high). 
For these datasets who have lower accuracy we also see that as the total available budget drops (Figures \ref{fig:medslr_experiments}(b) and (e)) classification performances degrade significantly, as when we run out of money we use only machines to classify.
We do not show here the stochastic policy because it is dominated by the fixed switch point (given that we determine a proportion of budget to learning, it is more efficient to learn first rather than later).

\section{Conclusion}  

The results show that even a small fraction of budget invested in ML training leads to improvement in cost and accuracy with respect to the state of the art screening algorithm available. Increasing the budget allocation to learning up to 50\% increases accuracy but also cost, and from that point on the improvements vary.
We observe that a difference with respect to crowd-only classification become more manifest as crowd accuracy grows over 0.7-0.8 as this leads to ``good'' training data. 
In summary, adding a sprinkle of ML to crowd classification and a sprinkle of crowd to machine classification leads to very high accuracy and better results in finite pool problems, provided that the budget vs pool size ratio is such that we can collect crowd votes rather than use ML only. 
While we tried to obtain diverse datasets, the applicability we claim is limited to screening problems with binary predicates, and to the HC algorithm adopted in the experiment.

\bibliographystyle{plain}
\bibliography{crowdsourcingFixed}

\section*{Supplementary Material}
In this supplementary material, we report F1 results
for the same experiments conducted in the paper. Note that while F1 score one of the most common choices for classification tasks to assess the accuracy of models, our multi-predicate classification problems preserve just little positive items, and requesters usually aim to find all relevant samples in the pool (for example, as in literature reviews). Therefore, optimizing classifiers towards high Recall as the loss of False Negatives is much higher than of False Positives (e. g., in liyerature reviews a False Positive paper means that an expert will have a look at it without sacrificing in resulting accuracy while a False Negative paper leads to an incomplete review).

\begin{figure*}[h]
	\centering
		\includegraphics[width=1.\textwidth]{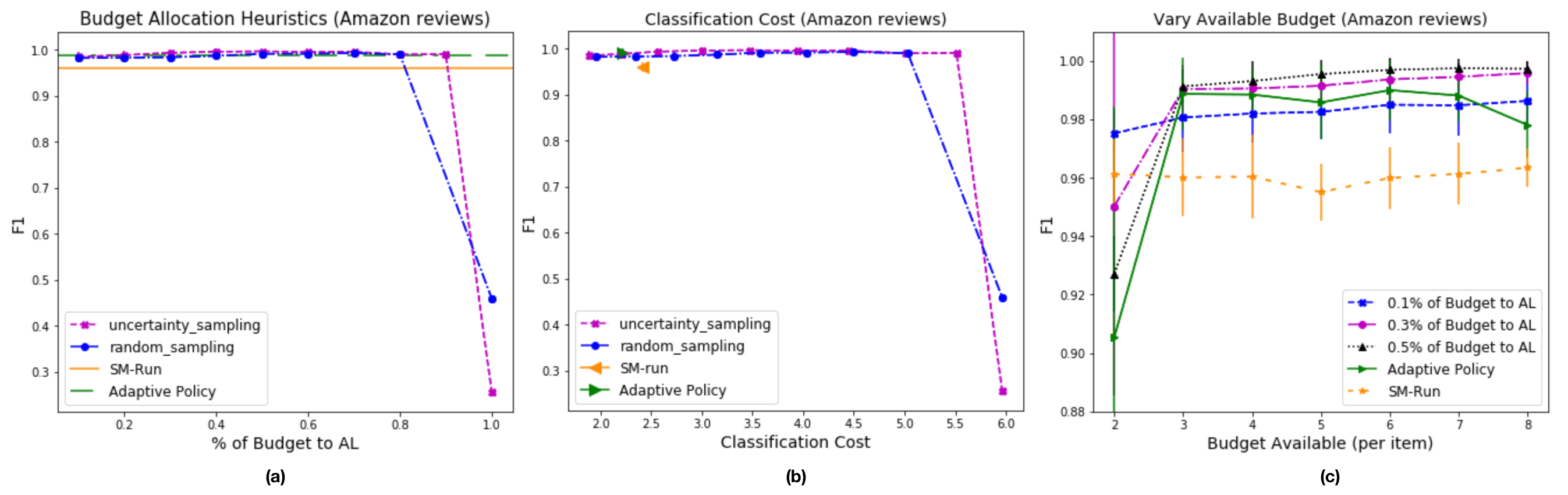}
    		\caption{Results on the Amazon reviews dataset with simulated crowd votes.}  
\label{fig:amazonresults2}
\end{figure*}

\begin{figure*}[h]
	\centering
		\includegraphics[width=1.\textwidth]{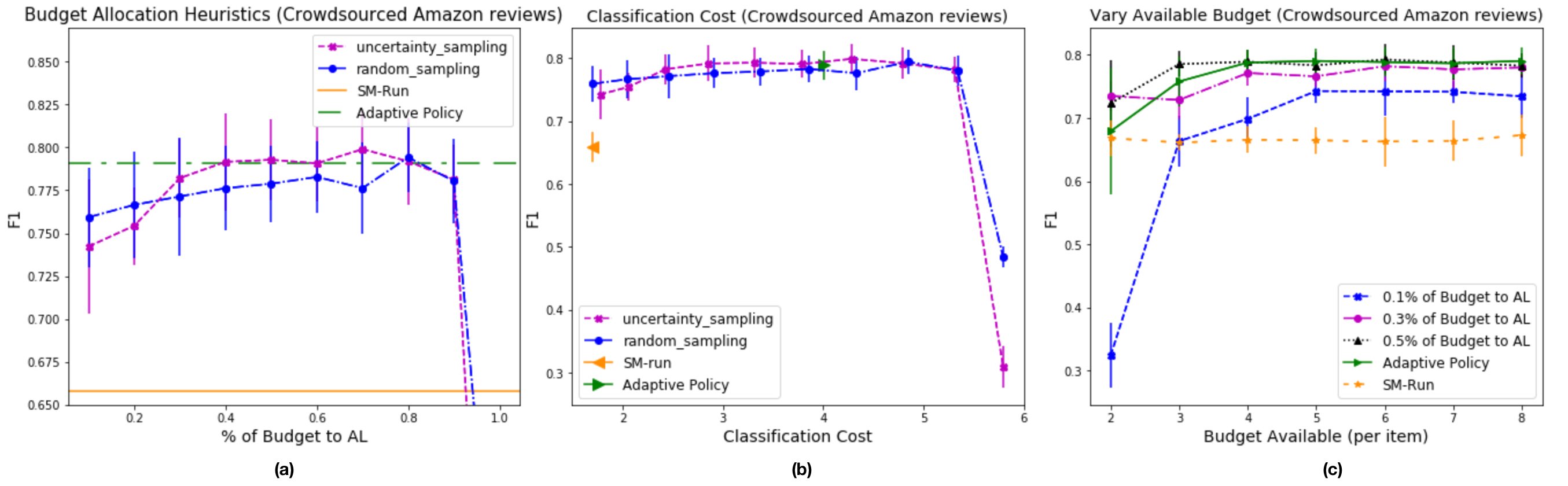}
    		\caption{Results on the Amazon reviews dataset with real crowd votes.}  
\label{fig:amazonresultsr2}
\end{figure*}

\begin{figure*}[h]
	\centering
		\includegraphics[width=1.\textwidth]{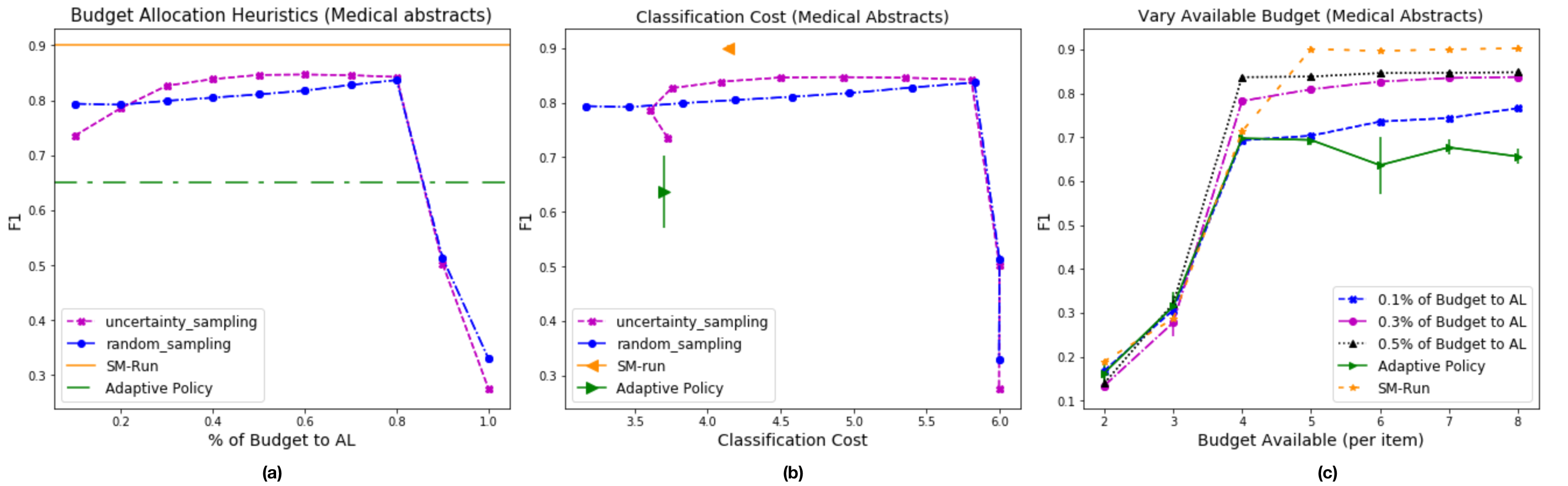}
    		\caption{Results on the MED Abstracts dataset.}  
\label{fig:med2}
\end{figure*}

\begin{figure*}[h]
	\centering
		\includegraphics[width=1.\textwidth]{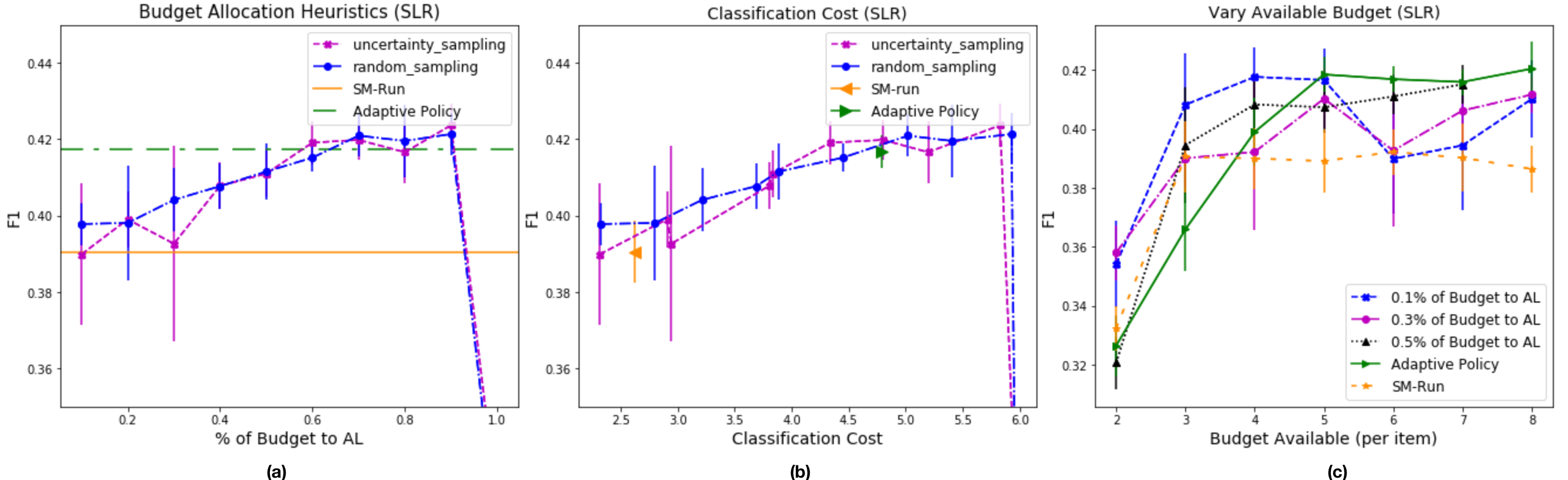}
    		\caption{Results on the SLR dataset.}  
\label{fig:slr2}
\end{figure*}

Figure~\ref{fig:amazonresults2}, \ref{fig:amazonresultsr2}, \ref{fig:med2}, \ref{fig:slr2} shows the F1 results on the Amazon reviews, MED Abstract, SLR  datasets with simulated and real crowd votes, where \textit{Classification Cost} is the actual cost spent per paper on average (in crowd votes), \textit{Budget Available} is average number of crowd votes we can spend per item, and \textit{\% of Budget to AL} represents different Fixed Point Switch policies, i.\,e., proportion of available budget to devote to ML training (that is, selecting the ML policy) and then switching to the crowd policy.

\subsection*{Amazon reviews with Simulated crowd votes}

\subsection*{Amazon reviews with Real crowd votes}

\newpage
\subsection*{Results on the MED Abstracts dataset.}

\subsection*{Results on the SLR dataset.}

\end{document}